# Feature transforms for image data augmentation

**Loris Nanni**[1,*], **Michelangelo Paci**[2], **Sheryl Brahnam**[3], **and Alessandra Lumini**[4]

1, Department of Information Engineering, University of Padua, Via Gradenigo 6, 35131 Padova, Italy; loris.nanni@unipd.it
2, BioMediTech, Faculty of Medicine and Health Technology, Tampere University, Arvo Ylpön katu 34, FI-33520, Tampere, Finland; michelangelo.paci@tuni.fi
3, Information Technology and Cybersecurity, Missouri State University, 901 S. National, Springfield MO, 65804, USA; SBrahnam@missouristate.edu
4, DISI, Università di Bologna, Via dell'università 50, 47521 Cesena, Italy; alessandra.lumini@unibo.it

\* Correspondence: loris.nanni@unipd.it; Tel.: + 39-373-85-35-801

**Abstract:** A problem with Convolutional Neural Networks (CNNs) is that they require large datasets to obtain adequate robustness; on small datasets, they are prone to overfitting. Many methods have been proposed to overcome this shortcoming with CNNs. In cases where additional samples cannot easily be collected, a common approach is to generate more data points from existing data using an augmentation technique. In image classification, many augmentation approaches utilize simple image manipulation algorithms. In this work, we propose some new methods for data augmentation based on several image transformations: the Fourier Transform (FT), the Radon Transform (RT), and the Discrete Cosine Transform (DCT). These and other data augmentation methods are considered in order to quantify their effectiveness in creating ensembles of neural networks. The novelty of this research is to consider different strategies for data augmentation to generate training sets from which to train several classifiers which are combined into an ensemble. Specifically, the idea is to create an ensemble based on a kind of bagging of the training set, where each model is trained on a different training set obtained by augmenting the original training set with different approaches. We build ensembles on the data level by adding images generated by combining fourteen augmentation approaches, with three based on FT, RT, and DCT, proposed here for the first time. Pretrained ResNet50 networks are finetuned on training sets that include images derived from each augmentation method. These networks and several fusions are evaluated and compared across eleven benchmarks. Results show that building ensembles on the data level by combining different data augmentation methods produce classifiers that not only compete competitively against the state-of-the-art but often surpass the best approaches reported in the literature.

## 1. Introduction

Deep learners, especially Convolutional Neural Networks (CNNs), are now the dominant classification paradigm in image classification, as witnessed by the plethora of articles in the literature that currently spotlight these networks. Learners like CNN are attractive, in part, because the architectures of these networks learn to extract salient features directly from samples, thus bypassing the need for human intervention in selecting the appropriate feature extraction method for the task at hand. These learned features have been shown to eclipse the power of handcrafted features chiefly because CNNs progressively downsample the spatial resolution of images while at the same time enlarging the depth of the feature maps.

Despite the strengths of CNNs, there are some significant drawbacks. Because the parameter size of CNNs is huge, these networks tend to overfit when trained on small datasets. Overfitting reduces the classifier's ability to generalize its learning so that it can correctly predict unseen samples. Researchers are now pressured to collect colossal datasets to accommodate the needs of deep learners, as exemplified by the ever-growing dataset ImageNet [1], which now contains over 14 million images classified into 1000 plus classes. In many domains, such as medical image analysis and bioinformatics (where samples might only number in the hundreds), collecting sufficient data for proper CNN training is prohibitively expensive and labor-intensive. This need for enormous datasets also requires that researchers have access to costly machines with considerable computational power.

Several solutions to the problem of overfitting that bypass the need for collecting more data have been proposed. Two of the most powerful techniques are transfer learning, where a given CNN architecture is pretrained on a massive dataset and provided to researchers and practitioners so that the network can be finetuned on smaller datasets, and 2) data augmentation, which adds new data points based on the original samples in a training set. Other methods unrelated to the techniques employed here include dropout [2], zero-shot/one-shot learning [3, 4], and batch normalization [2].

This study focuses on data augmentation since it has become a vital technology in fields where large datasets are difficult to procure [5-7]. Data augmentation methods aim at increasing the amount of training data by adding slightly modified copies of already existing data or newly created synthetic data from existing data. They act as a regularizer and helps reduce overfitting when training a machine learning model. Not only does data augmentation promote learning that leads to better CNN generalization, but it also fixes the problem of overfitting by adding and extracting information that is inherent within the training space. Data augmentation (DA) is a key element in the success of CNN models, as its use can lead to a faster convergence to solution and better prediction accuracy.

In the literature (see surveys [5-7]), most of the work focuses essentially on geometric transforms, statistical methods for color modification, and, recently, also on learned methods such as those based on GAN. To the best of our knowledge, this is the first paper that reports a very large study that deals with data augmentation approaches based on feature transform, testing them in several datasets spanning different applications.

Specifically, we focus on data augmentation techniques for image classification. These techniques can be divided into two broad types depending on whether the methods are based on basic image manipulations (such as translating and cropping) or on deep learning approaches [5]. The main object of this study is to evaluate the feasibility of building ensembles at the data level by adding augmented images generated using different sets of image manipulation methods, an approach that was taken in [8]. Unlike [8], however, this work performs a more exhaustive study building ensembles of augmentation methods by assessing over twice the number of techniques across eleven (versus only four) benchmark datasets.

The remainder of this paper is organized as follows: in Section 2, we review some of the best-performing data image manipulation approaches. In Section 3, novel data augmentation algorithms based on the radon transform (RT) [9], the discrete cosine transform (DCT), and the Fourier transform (FT) are proposed. As described more fully in Section 3, ensembles are built with pretrained ResNet50s finetuned on training sets composed of images taken from the original data and generated by an augmentation technique. These networks and their fusions are evaluated on the benchmarks described at the end of Section 3. In Section 4, we compare the performance of individual augmentation approaches and the ensembles built on them. The best ensemble reported in this work either exceeds the performance of the state-of-the-art in the literature or achieves similar performance on all the tested datasets. In Section 5, we conclude with a few suggestions for further research in this area.

The main contributions of this study can be summarized as follows:

- Presented is an extensive evaluation of common image manipulation methods used for data augmentation across eleven freely available and diverse benchmarks.
- Proposed are three new augmentation approaches utilizing RT, FT, and DCT transforms.
- Demonstrated is the value of building deep ensembles of classifiers on the data level by adding to the training sets images generated using different data augmentation approaches: the experimentally derived ensemble developed in this work is shown to achieve state-of-the-art performance on several benchmarks.
- Provided to the public at no charge is the MATLAB source code used in the experiments reported in this work (available at https://github.com/LorisNanni/Feature-transforms-for-image-data-augmentation).

## 2. Related Work

As mentioned in the Introduction, this study focuses on building ensembles with augmentation methods produced by the application of image manipulation methods. In [5], these methods are divided into the following groups based on: 1) simple geometric transformations, 2) randomly erasing and cutting, 3) mixing images, 4) kernel filters, and 5) color space transforms [5]. Most of these augmentation algorithms are easy to implement. Practitioners must be careful, however, when applying these image manipulations to a sample because it is possible to produce new images that no longer belong to the same class as the original. Flipping an image of the number six, for instance, would result in an image recognized as the number nine.

Flipping, especially along the horizontal axis, is one of the simplest and most popular geometric transforms for data augmentation [5], as are rotation (typically on the right or left axis in the range [1°, 359°]) and translation (where positional bias is avoided by shifting a sample up, down, left, and right) [5]. A problem with translation is that it can introduce undesirable noise [10]. Random cropping is another simple technique that reduces the size of new images, which is often needed to fit the input of a network. Augmented data can also be generated by merely substituting random values in an image, as extensively evaluated in [11]. In [12], the authors compared the performance of these simple augmentation techniques with each trained on AlexNet and assessed on two datasets, ImageNet and CIFAR10 [13]: rotation was found to perform better than translation, random cropping, and random values.

Random erasing [14] and cutting [15] occludes images; these methods model what occurs regularly in the real world, where objects are often only partially presented in the visual field. A review of the literature on augmentation methods based on this category can be found in [6]. Of particular interest is the method proposed in [14] that randomly erases an image with patches that vary in size. This method of partially erasing images was evaluated on ResNet architectures across three datasets: Fashion-MNIST, CIFAR10, and CIFAR100. Results showed consistent performance improvements.

Another simple method for constructing images is to mix them. A simple way to accomplish this task is to average the pixels between two or more images belonging to the same class [16]. Alternatively, images can be submitted to a transform, and the resulting components can be mixed, for example, by chaining, as in [17]. Masks can also be applied. In [16], the authors combined images using several image manipulation techniques: random images were flipped and cropped, then the RGB channel values for each pixel were averaged. Some nonlinear methods for mixing images were proposed in [18], and GANs were used in [18] to blend images.

Kernel filters can also be applied to create new images within a sample space. Filters are often used to sharpen or blur images. Filters, such as Gaussian blur, are applied by sliding an n × n window across the image. PatchShuffle, proposed in [19], randomly swaps matrix values in the filter window to make new images.

Novel color images can be created by means of color space transformations. A positive side-effect of this technique is the removal of illumination bias [5]. Transformations of color space can involve making a histogram of pixels in a color channel and applying different filters, much like those positioned over the lenses of cameras to alter the characteristics of the color space in a scene. Alternatively, color spaces can be converted into other color spaces. Care must be taken when transforming the color space, as it has been observed, for example, that changing an RGB image to a grayscale image can reduce the performance of a classifier [20]. New images can be produced by adding noise to color distributions or by jittering and adjusting the brightness, contrast, and saturation of samples [12, 21]. These color adjustments run the risk of removing valuable information. A review of color space transforms for image augmentation and a comparison of this type of image manipulation with geometric transforms is available in [22].

Not all data augmentation techniques take into account the entire training set. One popular technique in this vein is PCA jittering [12, 21-24], which produces new images by multiplying the PCA components by a small number. In [22], only the first component, which contains the most information, is jittered by being multiplied by a random number selected from a uniform distribution. Finally, in [23], an original image is transformed by PCA and DCT and jittered by adding noise to all components before reconstructing the image.

**3. Materials and methods**

*3.1. Proposed approach*

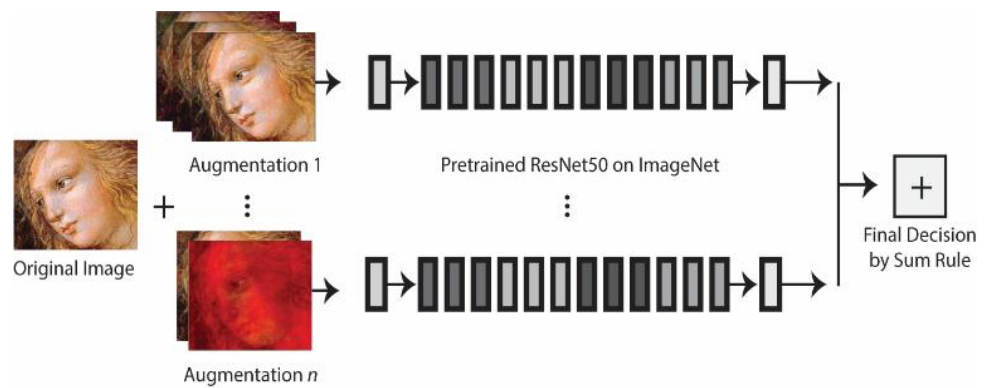

**Figure 1**. Ensembles are built on pretrained ResNet50s finetuned on training sets composed of original images combined with images generated by several data augmentation techniques.

The method proposed here for building ensembles of deep learners is illustrated in Figure 1. A given training set is augmented using $n = 14$ approaches, each detailed in Section 3.2. These new training sets are then used to finetune fourteen ResNet50s pretrained on ImageNet. In this work, each pre-trained ResNet50 is finetuned with a batch size of 30 and a learning rate of 0.001. In the testing phase, each unknown sample is classified by the 14 CNNs, and the resulting scores are fused by the sum rule.

*3.2. Data augmentation methods*

In this section, we describe the data augmentation sets (APPs1-14) explored in this study. APPs1-11 have been detailed in [8], so they will receive less attention. APPs12-14 are proposed here for the first time; these augmentations are explained more fully and illustrated in Figure 2 (resulting images) and Figure 3 (methods).

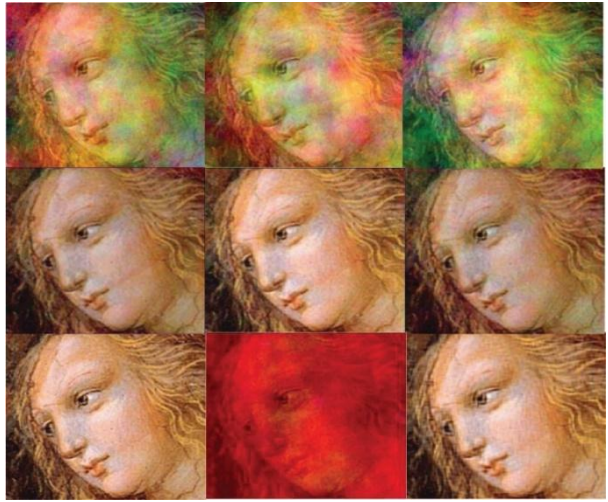

**Figure 2.** Examples of an original image (lower left-hand side) augmented on methods APP12 (top), APP13 (middle), and APP14 (bottom).

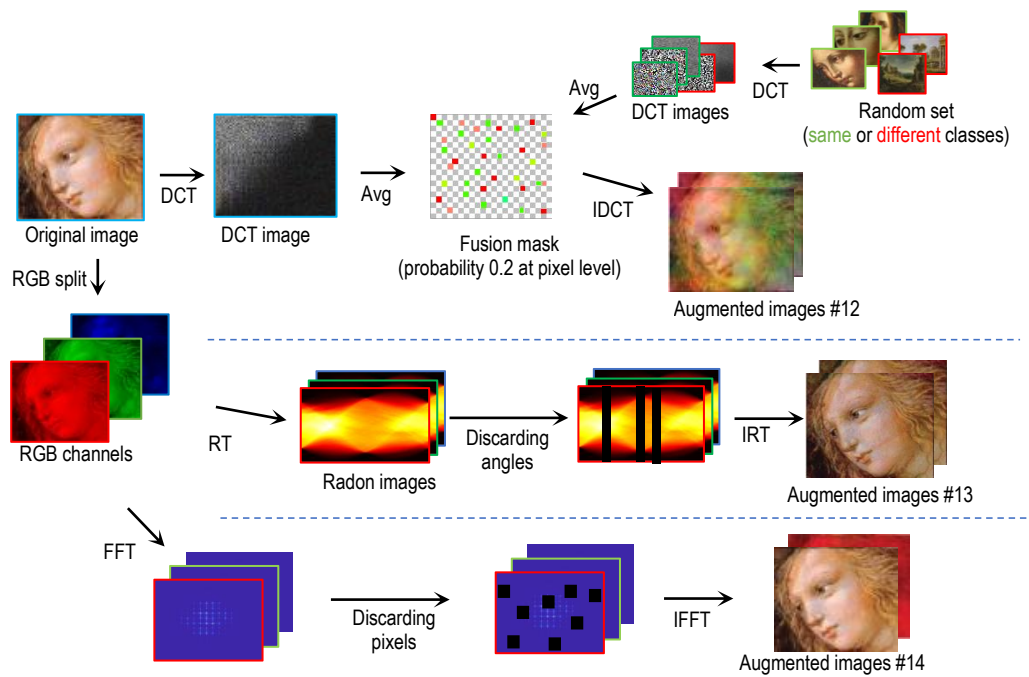

**Figure 3.** Schema for augmentation sets APP12-14 (designated here to save space as #12-#14).

The method proposed here for building ensembles of deep learners is illustrated in Figure 1. A given training set is augmented using $n = 14$ approaches, each detailed in Section 3.2. These new training sets are then used to finetune fourteen ResNet50s pretrained on the following training sets:

APP1 (number of new images generated: 3) takes a given image and randomly reflects it top-bottom and left-right for two new images. The third transform linearly scales the original image along both axes with two factors randomly extracted from the uniform distribution [1, 2].

APP2 (number of new images generated: 6) replicates App1 with three additional manipulations: image rotation (randomly extracted from [-10, 10] degrees), translation (along both axes with the value randomly sampled from the interval [0, 5] pixels), and shear (with vertical and horizontal angles randomly sampled in the interval [0, 30] degrees).

APP3 (number of new images generated: 4) replicates App2 without shear.

APP4 (number of new images generated: 3), proposed in [23], applies a transform based on PCA, where the PCA coefficients extracted from a given image are transformed 1) by randomly setting to zero (with a probability of 0.5) each element of the feature vector; 2) by adding noise to each component based on the standard deviation of the projected image; and 3) by selecting five images from the same class as the original image, computing the PCA vector, and randomly selecting components with a probability 0.05 from the original PCA vector and swapping them out with some of the corresponding components of the five other PCA vectors. The PCA inverse transform is applied to the three perturbed PCA vectors of the original image to produce the augmented images.

APP5 (number of new images generated: 3) applies the same perturbation method as those described in App4 using the DCT transform rather than PCA. With APP5, the DC coefficient is never changed.

APP6 (number of new images generated: 3) is applied to color images. The three images are constructed by altering contrast, sharpness, and color shifting. The contrast is altered by linearly scaling the original contrast of the image between the values a, the lowest value, and b, the highest value allowed for the augmented image. Every pixel in the original image outside this range is mapped to 0 if less than a or 255 if greater than b. Sharpness is altered by blurring the original with a Gaussian filter (variance = 1) and by subtracting the blurred image from the original. The Color is shifted with three integer shifts from three RGB filters. Each shift is then added to one of the three channels in the original image.

APP7 (number of new images generated: 7) is applied to color images. The first four augmented images are produced by altering the pixel colors in the original image using the MATLAB function jitterColorHSV with randomly selected values in the range of [0.05, 0.15] for hue, in the range [-0.4, -0.1] for hue saturation, in the range [-0.3, -0.1] for brightness, and in the range [1.2, 1.4] for contrast. Image five is generated with the MATLAB function imgaussfilt, where the standard deviation of the Gaussian filter randomly ranges in [1, 6]. Image six is generated by the MATLAB function imsharpen with sharpening equal to two and the radius of the Gaussian low-pass filter equal to one. Image seven applies the same color-shifting detailed in App6.

APP8 (number of new images generated: 2) is applied to color images and produces the two augmented images by randomly selecting a target image belonging to the same class as a given image followed by the application of two nonlinear mappings: RGB Histogram Specification and Stain Normalization using the Reinhard Method [25].

APP9 (number of new images generated: 6), proposed in [8], applies two elastic deformations: one MATLAB method that introduces distortions into the original image and an RGB adaptation of ElasticTransform from the computer vision tool Albumentations (available at https://albumentations.ai/ accessed 01/15/22). Both methods transform a given image by applying a randomly generated displacement field to its pixels by a value extracted from the standard uniform distribution in the range [-1,1] for the first method or in the range of [-1. +1] for the second. The resulting horizontal and vertical displacement fields are passed through three low-pass filters: 1) circular averaging filter, 2) rotationally symmetric Gaussian low-pass filter, and 3) rotationally symmetric Laplacian of Gaussian filter. For more details, see [8].

APP10 (number of new images generated: 3), proposed in [8], is based on DWT [26] (specifically, Daubechies wavelet db1 with one vanishing moment). DWT produces four matrices: the approximation coefficients (cA) and the horizontal, vertical, and diagonal coefficients (cH, cV, and cD, respectively). APP10 performs three perturbations on these matrices to generate three new images. In the first method, each element in the coefficient matrices is randomly selected, with a probability of 0.5, to be set to zero. In the second method, a constant is added to each element that is calculated by summing the standard deviation of the original image with a number randomly selected in the range [-0.5, 0.5]. In the third method, five images from the same class as the original image are randomly selected, and the DWT coefficient matrices are calculated for each one. Elements of the original cA, cH, cV, and cD matrices are then replaced, with a probability of 0.05, with values in the matrices of the other five images. The three augmented images are produced by applying the inverse DWT transform on the three sets of perturbed matrices.

APP11 (number of new images generated: 3): was first proposed in [8] and is based on the Constant-Q Transform (CQT) [27]. After calculating the CQT arrays of a given image, it undergoes the same perturbations to generate the three images as in APP10. The three augmented images are then produced by applying the inverse CQT transform on the perturbed CQT arrays.

APP12 (number of new images generated: 5) is a new method proposed here based on DCT and the random selection of other images. Three novel images are extracted from the same class as the original image and two from a different class. The original image and the five selected images are projected on the DCT space, with each element of the five images having a probability of 0.2 of being averaged with the original DCT element. Notice that this approach is cumulative, as can be observed in the following pseudo-code:

```
OriginalDCT=DCT(originalImage);
For selectedImage=1:5
  NewDCT=DCT(selectedImage);
  Sel = mask for averaging operation;
  OriginalDCT(Sel)=(OriginalDCT(Sel)+NewDCT(Sel))./2;
end;
NewImage=IDCT(OriginalDCT);
```

IDCT is applied at the end. Since there are three channels for each color image, these perturbations are applied to each channel independently.

APP13 (number of new images generated: 3) takes an original image and builds 3 augmented images, using the Radon transform, as can be observed in the following pseudo-code:

```
angles= randSel(160,[0:179]);
radonIm=Radon(originalImage,angles);
NewImage1=IRadon(radonIm,angles);

angles= [0:179];
radonIm=Radon(originalImage,angles);
columns= randSel(27,[0:179]);
radonIm(:,columns)=0
NewImage2=IRadon(radonIm,angles)

angles= randSel(160,[0:179]);
radonIm=Radon(originalImage,angles);
columns= randSel(27,[0:179]);
radonIm(:,columns)=0
NewImage3=IRadon(radonIm);
```

where randSel(num, range) randomly select num values in a range.

The first image is produced with the Radon Transform (RT), which projects the original image's intensity along a radial line oriented at a specific angle (angles with values between [0,179]), but, for the first image, twenty angles are randomly selected and discarded. Then the image is back-projected with the Inverse RT (IRT). The second image is generated in the same way as the first, but all angles are used to project the image with RT, after which 15% of the angle values (that is, the columns of the projected image) are set to zero before IRT is applied. The third image is like the first in that 20 angles are randomly selected and discarded in the projection step, and, like in the construction of the second image, only 15% of the angles (columns of the projected image) are set to zero before IRT is applied.

APP14 (number of new images generated: 2) The first image is generated using the Fast Fourier Transform (FFT). After FFT is applied, 50% of the coefficients are randomly set to zero before performing the inverse FFT. The second image is built by applying DCT; next, a square low-frequency filter (size 40×40) is applied on the DCT image before performing the inverse DCT:

```
fftIm=FFT(originalImage);
rndMask=randomMask(originalImage,0.5);
fftIm(rndMask)=0;
NewImage1=IFFT(D fftIm);

dctIm=DCT(originalImage);
dctIm (i>40,j>40,:)=0;
NewImage2=IDCT(dctIm);
```

where randomMask(image, prob) returns a random pixel mask which is of the same size as the image and prob is the probability of each pixel to be 1.

*3.3. Datasets*

In this work, ensembles of augmentation methods are tested and compared with the literature on eleven benchmark datasets for image classification (see Table 1).

**Table 1.** Description of the eleven datasets used in this study.

| Short Name | Full Name | #Classes | #Samples | Image Size | Protocol | Ref |
|---|---|---|---|---|---|---|
| VIR | Virus | 15 | 1500 | 41×41 | 10CV | [28] |
| BARK | Bark | 23 | 23000 | ~1600×3800 | 5CV | [29] |
| GRAV | Gravity | 22 | 8583 | 470×570 | Tr-Te | [30] |
| POR | Portraits | 6 | 927 | From 80×80 to 2700×2700 | 10CV | [31] |
| PBC | Peripheral blood cell classification | 8 | 17092 | 360x363 | Tr-Te | [32] |
| HE | 2D HELA | 10 | 862 | 512×382 | 5CV | [33] |
| MA | Muscle aging | 4 | 237 | 1600×1200 | 5CV | [34] |
| BG | Breast grading carcinoma | 3 | 300 | 1280×960 | 5CV | [35] |
| LAR | Laryngeal dataset | 4 | 1320 | 1280×960 | Tr-Te | [36] |
| Triz | Gastric lesion types | 4 | 574 | 352×240 | 10CV | [37] |
| END | Histopathological endometrium images | 4 | 3502 | 640×480 | Tr-Te | [38] |

In Table 1, the following information is reported for each dataset: a short name, the original dataset name (if provided in the reference), the number of classes and samples, the size(s) of the images, the testing protocol, and the original reference. The abbreviations for the testing protocols in Table 1 are detailed as follows:
- 5CV,10CV represents 5-fold and 10-fold cross-validation.
- Tr-Te represents a dataset that is pre-divided into training and testing sets. For LAR, a three-fold division is provided by the authors. For PBC, the official protocol specifies that 88% of the images be included in the training set and 12% in the test set, with both sets maintaining the same sample per class ratio as in the original dataset. END includes a training set of 3302 images and an external validation set of 200 images.

The performance indicator typically reported on these datasets is accuracy, which measures the rate of correct classifications. For the GRAV dataset, four different views are extracted at different durations from each glitch/image; therefore, the final score is obtained by combining the four classification scores via the average rule. Validation of the superiority of one method over the others is provided by the Wilcoxon signed rank test [39].

**4. Experimental results**

We start our experiments by comparing the performance of the augmentation sets with ResNet50 (see Table 2), along with the results of these approaches on the following ensembles:
- EnsDA_A: this is the fusion by sum rule among all the ResNet50 trained using App1-11; each ResNet50 is trained with a different data augmentation approach. The data augmentation methods based on color spaces (App6-8) are not reported on VIR, HE, and MA since they are gray-level images.
- EnsDA_B: this fusion is the same as EnsDA_A except for the addition of ResNet50s trained with the new augmentation methods App12-14.
- EnsDA_C: this is the fusion by sum rule among those methods not based on feature transforms. Each approach is iterated twice (three times for datasets VIR, HE, and MA since they are gray-level images; they are trained three times so that the size of the ensemble EnsDA_C is similar to EnsDA_B).
- EnsBase(X): this is a baseline ensemble intended to compare/validate the performance of EnsDA_* (i.e., 1-3 above); EnsBase(X) combines (via sum rule) X ResNet50 networks trained separately on App3, which produces the best average performance compared with all the other data augmentation sets.

**Table 2.** Performance accuracy (in %) of data augmentation sets (APP1-14) and defaults.

| DataAUG | VIR | HE | MA | BG | LAR | POR | Bark | Grav | TriZ | END | PBC |
|---|---|---|---|---|---|---|---|---|---|---|---|
| NoDA | 85.53 | 95.93 | 95.83 | 92.67 | 94.77 | 86.29 | 87.48 | 97.66 | 98.97 | 55.50 | 98.98 |
| App1 | 87.00 | 95.12 | 95.00 | 93.00 | 92.95 | 87.05 | 89.60 | 97.83 | 99.13 | 56.00 | 98.93 |
| App2 | 86.87 | 96.63 | 95.83 | **94.00** | 95.08 | 85.97 | 90.17 | 98.08 | 99.13 | 51.50 | 99.08 |
| App3 | 87.80 | 95.12 | 95.00 | **94.00** | 94.55 | 87.05 | 89.45 | 97.99 | 98.96 | 56.50 | 98.88 |
| App4 | 86.33 | 95.23 | 93.33 | 92.33 | 94.62 | 84.90 | 87.91 | 97.74 | 98.08 | 75.00 | 98.74 |
| App5 | 86.00 | 95.35 | 91.25 | 91.33 | 95.45 | 86.41 | 87.61 | 97.83 | 98.43 | 77.50 | 98.74 |
| App6 | ** | ** | ** | 92.33 | 94.39 | 87.37 | 88.63 | 98.08 | 98.43 | 75.00 | 98.78 |
| App7 | ** | ** | ** | 93.33 | 95.08 | 88.13 | 89.28 | 97.99 | 98.61 | 81.00 | 98.98 |
| App8 | ** | ** | ** | 90.67 | 94.70 | 86.06 | 87.29 | 97.74 | 98.26 | 80.50 | 98.35 |
| App9 | 85.67 | 95.58 | 94.17 | 91.67 | 95.15 | 86.19 | 88.86 | 98.24 | 98.95 | 69.50 | 98.93 |
| App10 | 84.20 | 95.81 | 91.25 | 88.67 | 93.64 | 85.10 | 86.39 | **98.41** | 99.31 | 62.00 | 98.64 |
| App11 | 85.47 | 95.35 | 91.25 | 92.67 | 95.98 | 86.71 | 89.20 | 97.91 | **99.48** | 80.00 | 98.69 |
| App12 | 86.73 | 95.23 | 91.67 | 90.33 | 95.45 | 85.63 | 86.81 | 97.49 | 97.90 | 71.00 | 98.78 |
| App13 | 86.20 | 94.77 | 92.92 | 91.67 | 95.15 | 85.76 | 88.40 | 98.08 | 98.08 | 75.00 | 98.69 |
| App14 | 85.27 | 95.47 | 91.25 | 93.33 | 93.71 | 87.26 | 87.84 | 97.91 | 98.43 | **84.00** | 98.83 |
| DivAUG [40] | 86.07 | 95.23 | 90.83 | 89.33 | 96.29 | 86.52 | *** | *** | *** | *** | *** |
| EnsDA_A | 90.00 | 96.51 | 97.08 | **94.00** | 96.29 | 89.21 | 91.27 | 98.33 | 99.13 | 76.00 | 98.98 |
| EnsDA_B | **90.20** | **96.63** | 97.08 | **94.00** | 96.14 | 89.96 | 91.00 | 98.24 | 99.13 | 77.50 | 99.08 |
| EnsDA_C | 89.33 | 96.51 | 97.08 | 93.67 | **96.74** | 90.07 | 91.38 | 98.33 | 99.13 | 71.50 | **99.12** |
| Ens_Base(14) | 89.73 | 96.40 | **97.50** | 93.67 | 96.14 | 88.02 | 90.66 | 98.08 | 98.78 | 50.50 | 98.88 |
| Ens_Base(11) | 89.73 | 96.28 | **97.50** | 93.67 | 95.91 | 87.58 | 90.67 | 98.16 | 98.78 | 50.00 | 98.74 |
| Ens_Base(5) | 89.60 | 95.93 | 96.67 | 93.33 | 96.14 | 87.48 | 90.66 | 97.99 | 98.78 | 49.00 | 98.78 |

**Augmentations that work on color images were not run on gray-level images. *** Computation time exceeded resources.

The label NoDA in the first row of Table 2 is a stand-alone ResNet50 trained without data augmentation.

Several conclusions can be drawn by examining Table 2:
- The best augmentation set varies with each dataset: in some, the best approach is a feature transform (see GRAV, Triz, and END); in others, a color-based method (see POR), and for some, the best performance is obtained using affine transformations.
- Considering a stand-alone CNN, in some datasets (see PBC), the performance of NoDA is similar to that of the best App augmentation sets.
- In general, though, the ensembles strongly boost the performance of NoDA: both Ens_Base(11) and Ens_Base(14) outperform NoDA with a p-value of 0.1, and all the EnsDA_* outperform NoDA with a p-value of 0.001. Across all the datasets, the EnsDA_* ensembles obtain an accuracy higher than or equal to that obtained by NoDA.
- EnsDA_C outperforms Ens_Base(14) with a p-value of 0.1. Both EnsDA_A and EnsDA_B (which include the augmentation methods based on the feature transform approaches) outperform Ens_Base(14) with a p-value of 0.05. Among the different tested ensembles, our suggested approach is EnsDA_B since it obtains the highest average performance among the EnsDA_* (EnsDA_A: 93.34%, EnsDA_B: 93.54%, EnsDA_C: 92.98%, EnsBase(14): 90.76%).

The approach proposed in [40] selects only a subset of images from a larger set of built images. Here, as a base, we use a large dataset made up of the images created by all the methods belonging to EnsDA_B, but it produced no improvement in performance compared to EnsDA_B; for this reason, and for the sake of reducing computation time, it was tested only on a subset of all the datasets.

A second experiment was performed to confirm the previous results on a different architecture. In Table 3, the same ensembles reported in Table 2 are evaluated using mobileNetV2 [41] instead of ResNet50. MobileNetv2 is a lightweight architecture that produces results comparable to heavy architectures using far less computational resources.

**Table 3.** Performance accuracy (in %) using MobileNetv2 as a model.

| MobileNet | VIR | HE | MA | BG | LAR | POR | Bark | Grav | TriZ | END | PBC |
|---|---|---|---|---|---|---|---|---|---|---|---|
| EnsDA_A | **85.27** | 96.16 | 95.83 | 93.00 | **96.21** | 88.56 | 91.20 | 98.16 | 98.25 | 86.00 | 99.17 |
| EnsDA_B | 84.47 | 95.47 | 96.25 | **93.33** | 95.98 | 88.55 | 90.95 | **98.24** | 98.26 | **87.00** | 99.22 |
| EnsDA_C | 74.73 | **96.63** | 95.00 | **93.33** | 95.76 | 87.91 | **91.56** | 98.16 | 98.25 | 86.50 | 99.22 |
| Ens_Base(14) | 47.60 | 95.00 | **97.50** | 92.67 | 95.30 | 85.96 | 91.04 | 98.16 | **98.26** | 83.00 | **99.37** |
| Ens_Base(11) | 47.60 | 95.00 | 97.08 | 92.67 | 94.85 | 85.86 | 91.10 | 98.16 | **98.26** | 82.50 | 99.27 |
| Ens_Base(5) | 47.53 | 95.47 | 96.25 | 93.00 | 94.32 | 85.64 | 90.90 | 97.99 | **98.26** | 82.50 | 99.27 |

The results in Table 3 substantially confirm conclusions reported in Table 2, *viz.*, the proposed approaches for data augmentation are valid methods for increasing diversity among classifiers and designing high-performing ensembles: EnsDA_* outperform EnsBase(14) with a p-value of 0.1.

Even if accuracy is probably the most used performance indicator for classification problems, it is not the most suitable for comparing classifiers. The area under the curve (AUC) is preferred as a standard measure in tests of predictive modeling performance. The AUC is an estimate of the probability that a classifier will rank a randomly chosen positive instance higher than a randomly chosen negative instance. In this work, we use the one's complement of AUC, i.e., the error under the ROC curve (EUC): EUC=1-AUC. Thus, in Tables 4 and 5, the performance of the proposed approaches in terms of EUC is reported. Because EUC is an indicator for binary classifiers, in multiclass problems, the average value of one-versus-all EUC is used (the rocmetrics MATLAB function has been employed). The results in Tables 4 and 5 largely reflect the trend of accuracy, namely, that the proposed ensembles based on data augmentation outperform both base ensembles and stand-alone approaches.

**Table 4.** EUC (1-AUC) using ResNet50 as a model.

| ResNet50 | VIR | HE | MA | BG | LAR | POR | Bark | Grav | TriZ | END | PBC |
|---|---|---|---|---|---|---|---|---|---|---|---|
| EnsDA_A | 1.37 | 0.24 | 0.27 | **2.39** | 0.12 | 1.75 | **1.38** | 0.23 | 0.04 | 10.73 | **0.01** |
| EnsDA_B | **1.33** | 0.23 | 0.27 | 2.50 | **0.11** | 1.68 | 1.40 | 0.22 | 0.05 | **9.24** | **0.01** |
| EnsDA_C | 1.42 | 0.20 | **0.11** | 3.14 | 0.17 | 1.71 | 1.37 | **0.30** | 0.07 | 9.89 | **0.01** |
| Ens_Base(14) | 1.36 | 0.23 | 0.21 | 2.42 | 0.12 | 2.45 | 1.56 | 0.31 | 0.13 | 20.48 | 0.03 |
| Ens_Base(11) | 1.37 | 0.27 | 0.23 | 2.38 | 0.15 | 2.48 | 1.55 | 0.29 | 0.12 | 20.94 | 0.03 |

**Table 5.** EUC using MobileNetv2 as a model.

| MobileNet | VIR | HE | MA | BG | LAR | POR | Bark | Grav | TriZ | END | PBC |
|---|---|---|---|---|---|---|---|---|---|---|---|
| EnsDA_A | 2.35 | **0.17** | 0.20 | 3.06 | 0.24 | 2.04 | 1.26 | 0.31 | **0.06** | 4.79 | 0.02 |
| EnsDA_B | **2.24** | **0.17** | 0.16 | 2.77 | 0.23 | **1.96** | **1.30** | 0.29 | 0.07 | **4.63** | 0.02 |
| EnsDA_C | 3.71 | 0.22 | **0.14** | 3.14 | 0.20 | 2.10 | 1.24 | 0.36 | 0.06 | 4.73 | **0.01** |
| Ens_Base(14) | 17.46 | 0.24 | **0.19** | 2.32 | 0.28 | 2.74 | 1.34 | 0.34 | 0.17 | 6.72 | **0.01** |
| Ens_Base(11) | 17.56 | 0.24 | 0.22 | 2.36 | 0.32 | 2.75 | 1.36 | 0.33 | 0.17 | 6.96 | **0.01** |

The results reported in Tables 4/5 substantially confirm previous conclusions:

EnsDA_A and EnsDA_B outperforms (considering EUC) EnsBase(14) with a p-value of 0.1 in both the tested network topologies (MobileNetV2 and ResNet50). EnsDA_B obtains an average performance better than that obtained by EnsDA_A.

In Table 6, our best ensemble is compared with the best methods reported in the literature on the same datasets. As can be observed, our proposed method obtains state-of-the-art or similar performance. Note that the performance indicator is the F1-measure with the LAR dataset because that is the measure that is reported most commonly in the literature for this dataset.

**Table 6.** Performance as a measure of accuracy (in %) compared with the best in the literature.

| DATASET | ResNet50 EnsDA_B | MobileNetV2 EnsDA_B | [42] | [43] | [44] | [45] | [46] | [28] | [45] | [47] |
|---|---|---|---|---|---|---|---|---|---|---|
| VIR | 90.20 | 84.47 | 89.60 | 89.47 | 89.00 | 88.00 | 87.27 | 87.00*** | 86.20 | 85.70 |
| LAR** | **EnsDA_B** 96.09 | **EnsDA_B** 95.98 | [48] 95.2 | [36] 92.0 | | | | | | |
| POR | **EnsDA_B** 89.96 | **EnsDA_B** 88.55 | [31] 90.08 | | | | | | | |
| BARK | **EnsDA_B** 91.00 | **EnsDA_B** 90.95 | [49] 48.90 | [50] 85.00 | [51] 90.40 | [29] 85.00 | | | | |
| PBC | **EnsDA_B** 99.08 | **EnsDA_B** 99.22 | [52] 99.30 | [53] 97.94 | | | | | | |
| GRAV | **EnsDA_B** 98.24 | **EnsDA_B** 98.24 | [30] 98.21 | | | | | | | |
| Triz | **EnsDA_B** 99.13 | **EnsDA_B** 98.26 | [37]**** 87.00 | | | | | | | |
| END | **EnsDA_B** 77.50 | **EnsDA_B** 87.00 | [38] 76.91 | | | | | | | |
| HE | **EnsDA_B** 96.63 | **EnsDA_B** 95.47 | [54] 98.30 | [55] 94.40 | [35] 84.00 | [56] 68.30 | | | | |
| MA | **EnsDA_B** 97.08 | **EnsDA_B** 96.25 | [54] 97.90 | [57] 53.00 | [56] 89.60 | | | | | |
| BG | **EnsDA_B** | **EnsDA_B** | [35] | [48] | | | | | | |

| | | | |
|---|---|---|---|
| 94.00 | 93.33 | 96.30 | 95.00 |

\*\* On LAR, F1 is the performance measure. \*\*\* The method in [28] combines descriptors based on both object scale and fixed scale images. \*\*\*\* Only handcrafted features are used.

In figure 4, we evaluate the disagreement of predictions on different CNNs to evaluate their degree of diversity [58]. The cosine similarity among scores is calculated for eight networks: the first five are networks trained separately on App3 (we used POR dataset for this experiment). In contrast, the last three are trained on an augmented dataset by APP12, APP13, and APP14, respectively. As can be observed from figure 4, the dissimilarity among scores is maximized in the last three rows/columns, proving the higher diversity among classifiers.

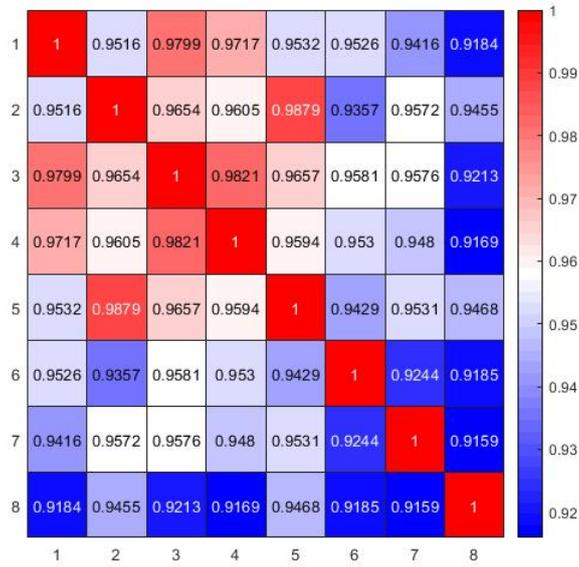

**Figure 4.** Cosine Similarity between scores of eight ResNet50: 1-5 trained on the original training set and 6-8 trained on an augmented training set.

## 5. Conclusion

In this study, we compare combinations of pretrained ResNet50s finetuned on training sets with the addition of some of the best-performing image manipulation methods for generating new images. The performance of these networks and their fusions were compared across eleven benchmarks representing diverse image classification tasks.

This study shows that constructing ensembles of deep learners on the data level by adding images generated by different data augmentation techniques increases the robust-ness of CNNs. Given the breadth in the diversity of the selected benchmarks, the approach taken here for building CNN ensembles should work on most image problems.

**Acknowledgments:** The authors wish to acknowledge the NVIDIA Corporation for supporting this research with the donation of a Titan Xp GPU and the TCSC–Tampere Center for Scientific Computing for generous computational resources.

**Funding**     No funds, grants, or other support was received.

**Data availability** The datasets generated during and/or analysed during the current study are available from the corresponding author on reasonable request.

**Declarations**

**Conflicts of interest**     The authors declare that there is no any conflict of interest

# References

[1]     J. Deng, W. Dong, R. Socher, L. Li, K. Li, and L. Fei-Fei, "ImageNet: A large-scale hierarchical image database," in *CVPR*, Miami, FL, 2009: IEEE, pp. 248-255.

[2]     V. Shirke, R. Walika, and L. Tambade, "Drop : A Simple Way to Prevent Neural Network by Overfitting," *International Journal of Research in Engineering, Science and Management,* vol. 1, no. 9, pp. 2581-5782, 2018.

[3]     M. Palatucci, D. A. Pomerleau, G. E. Hinton, and T. M. Mitchell, "Zero-shot Learning with Semantic Output Codes," in *Neural Information Processing Systems (NIPS)*, Vancouver, British Columbia, Canada, 2009, vol. 22.

[4]     Y. Xian, C. H. Lampert, B. Schiele, and Z. Akata, "Zero-Shot Learning-A Comprehensive Evaluation of the Good, the Bad and the Ugly," *IEEE Transactions on Pattern Analysis and Machine Intelligence,* vol. 41, no. 9, pp. 2251-2265, 2019.

[5]     C. Shorten and T. M. Khoshgoftaar, "A survey on Image Data Augmentation for Deep Learning," *Journal of Big Data,* vol. 6, no. 60, pp. 1-48, 2019.

[6]     H. Naveed, "Survey: Image Mixing and Deleting for Data Augmentation," *ArXiv,* vol. abs/2106.07085, 2021.

[7]     C. Khosla and B. S. Saini, "Enhancing Performance of Deep Learning Models with different Data Augmentation Techniques: A Survey," in *International Conference on Intelligent Engineering and Management (ICIEM)*, 2020, pp. 79-85, doi: 10.1109/ICIEM48762.2020.9160048.

[8]     L. Nanni, M. Paci, S. Brahnam, and A. Lumini, "Comparison of Different Image Data Augmentation Approaches," *Journal of Imaging,* vol. 7, no. 12, p. 254, 2021. [Online]. Available: https://www.mdpi.com/2313-433X/7/12/254.

[9]     R. N. Bracewell, *Two-dimensional imaging*. Prentice-Hall, Inc., 1995.

[10]    A. Mikołajczyk and M. Grochowski, "Data augmentation for improving deep learning in image classification problem," in *2018 International Interdisciplinary PhD Workshop (IIPhDW)*, 9-12 May 2018, pp. 117-122, doi: 10.1109/IIPHDW.2018.8388338.

[11]    F. J. Moreno-Barea, F. Strazzera, J. M. Jerez, D. Urda, and L. Franco, "Forward Noise Adjustment Scheme for Data Augmentation," *2018 IEEE Symposium Series on Computational Intelligence (SSCI),* pp. 728-734, 2018.

[12]    J. Shijie, W. Ping, J. Peiyi, and H. Siping, "Research on data augmentation for image classification based on convolution neural networks," in *Chinese Automation Congress (CAC) 2017*, Jinan, CN, 2017, pp. 4165-4170.

[13]    A. Krizhevsky, "Learning Multiple Layers of Features from Tiny Images," University of Toronto, 2009. [Online]. Available: https://www.cs.toronto.edu/~kriz/learning-features-2009-TR.pdf

[14]    Z. Zhong, L. Zheng, G. Kang, S. Li, and Y. Yang, "Random Erasing Data Augmentation," in *AAAI conference on artificial intelligence*, New York, 2020, vol. 34, no. 7, pp. 13001-13008.

[15]    T. Devries and G. W. Taylor, "Improved Regularization of Convolutional Neural Networks with Cutout," *ArXiv,* vol. abs/1708.04552, 2017.

[16]    H. Inoue, "Data Augmentation by Pairing Samples for Images Classification," *ArXiv,* vol. abs/1801.02929, 2018.

[17]    D. Hendrycks, N. Mu, E. D. Cubuk, B. Zoph, J. Gilmer, and B. Lakshminarayanan, "AugMix: A Simple Data Processing Method to Improve Robustness and Uncertainty," *ArXiv,* vol. abs/1912.02781, 2020.

[18]    D. Liang, F. Yang, T. Zhang, and P. Yang, "Understanding Mixup Training Methods," *IEEE Access,* vol. 6, pp. 58774-58783, 2018.

[19] G. Kang, X. Dong, L. Zheng, and Y. Yang, "PatchShuffle Regularization," *ArXiv,* vol. abs/1707.07103, 2017.

[20] K. Chatfield, K. Simonyan, A. Vedaldi, and A. Zisserman, "Return of the Devil in the Details: Delving Deep into Convolutional Nets," in *Proceedings British Machine Vision Conference*, University of Nottingham, Britian, 2014, doi: doi.org/10.5244/C.28.6.

[21] A. Krizhevsky, I. Sutskever, and G. E. Hinton, "COPY ImageNet classification with deep convolutional neural networks," in *Advances in Neural Information Processing Systems*, P. L. Bartlett, F. C. N. Pereira, C. J. C. Burges, L. Bottou, and K. Q. Weinberger Eds. Lake Tahoe, NV: Curran Associates, Inc., 2012, pp. 1106-1114

[22] L. Taylor and G. Nitschke, "Improving Deep Learning with Generic Data Augmentation," *2018 IEEE Symposium Series on Computational Intelligence (SSCI),* pp. 1542-1547, 2018.

[23] L. Nanni, S. Brahnam, S. Ghidoni, and G. Maguolo, "General Purpose (GenP) Bioimage Ensemble of Handcrafted and Learned Features with Data Augmentation," *ArXiv,* vol. abs/1904.08084, 2019.

[24] J. Nalepa, M. Myller, and M. Kawulok, "Training- and Test-Time Data Augmentation for Hyperspectral Image Segmentation," *IEEE Geoscience and Remote Sensing Letters,* vol. 17, pp. 292-296, 2020.

[25] A. M. Khan, N. Rajpoot, D. Treanor, and D. Magee, "A Nonlinear Mapping Approach to Stain Normalization in Digital Histopathology Images Using Image-Specific Color Deconvolution," *IEEE Transactions on Biomedical Engineering,* vol. 61, no. 6, pp. 1729-1738, 2014, doi: 10.1109/TBME.2014.2303294.

[26] D. Gupta and S. Choubey, "Discrete Wavelet Transform for Image Processing," *International Journal of Emerging Technology and Advanced Engineering,* vol. 4, no. 3, pp. 598-602, 2014.

[27] G. Angelo, Velasco, N. Holighaus, M. Dörfler, and T. Grill, "Constructing an invertible constant-q transform with nonstationary gabor frames," in *14th International Conference on Digital Audio Effects (DAFx 11)*, Paris, France, 2011, p. 33.

[28] G. Kylberg, M. Uppström, and I.-M. Sintorn, "Virus texture analysis using local binary patterns and radial density profiles," in *18th Iberoamerican Congress on Pattern Recognition (CIARP),* Havana, Cuba, S. Martin and S.-W. Kim, Eds., 2011, pp. 573-580.

[29] M. Carpentier, P. Giguère, and J. Gaudreault, "Tree Species Identification from Bark Images Using Convolutional Neural Networks," *2018 IEEE/RSJ International Conference on Intelligent Robots and Systems (IROS),* pp. 1075-1081, 2018.

[30] S. Bahaadini *et al.*, "Machine learning for Gravity Spy: Glitch classification and dataset," *Inf. Sci.,* vol. 444, no. May, pp. 172-186, 2018.

[31] S. Liu, J. Yang, S. S. Agaian, and C. Yuan, "Novel features for art movement classification of portrait paintings," *Image and Vision Computing,* vol. 108, p. 104121, 2021/04/01/ 2021, doi: https://doi.org/10.1016/j.imavis.2021.104121.

[32] A. Acevedo, A. Merino, S. Alférez, Á. Molina, L. Boldú, and J. Rodellar, "A dataset of microscopic peripheral blood cell images for development of automatic recognition systems," *Data in Brief,* vol. 30, p. 105474, 2020/06/01/ 2020, doi: https://doi.org/10.1016/j.dib.2020.105474.

[33] M. V. Boland and R. F. Murphy, "A neural network classifier capable of recognizing the patterns of all major subcellular structures in fluorescence microscope images of HeLa cells," *BioInformatics,* vol. 17, no. 12, pp. 1213-223, 2001.

[34] L. Shamir, N. V. Orlov, D. M. Eckley, and I. Goldberg, " IICBU 2008: a proposed benchmark suite for biological image analysis," *Medical & Biological Engineering & Computing,* vol. 46, no. 9, pp. 943–947, 2008.

[35] K. Dimitropoulos, P. Barmpoutis, C. Zioga, A. Kamas, K. Patsiaoura, and N. Grammalidis, "Grading of invasive breast carcinoma through Grassmannian VLAD encoding," *PLoS ONE,* vol. 12, pp. 1–18, 2017, doi: doi:10.1371/journal.pone.0185110.

[36] S. Moccia *et al.*, "Confident texture-based laryngeal tissue classification for early stage diagnosis support," *Journal of Medical Imaging (Bellingham),* vol. 4, no. 3, p. 34502, 2017.

[37] R. Zhao *et al.*, "TriZ-a rotation-tolerant image feature and its application in endoscope-based disease diagnosis," *Computers in biology and medicine,* vol. 99, pp. 182-190, 2018.

[38] H. Sun, X. Zeng, T. Xu, G. Peng, and Y. Ma, "Computer-Aided Diagnosis in Histopathological Images of the Endometrium Using a Convolutional Neural Network and Attention Mechanisms," *IEEE Journal of Biomedical and Health Informatics,* vol. 24, no. 6, pp. 1664-1676, 2020, doi: 10.1109/JBHI.2019.2944977.

[39] J. Demšar, "Statistical comparisons of classifiers over multiple data sets," *Journal of Machine Learning Research,* vol. 7 pp. 1-30, 2006.

[40] Z. Liu, H. Jin, T.-H. Wang, K. Zhou, and X. Hu, "DivAug: Plug-in Automated Data Augmentation with Explicit Diversity Maximization," in *IEEE/CVF International Conference on Computer Vision*, Virtual, 2021, pp. 4762-4770.

[41] M. Sandler, A. Howard, M. Zhu, A. Zhmoginov, and L. Chen, "MobileNetV2: Inverted Residuals and Linear Bottlenecks," in *2018 IEEE/CVF Conference on Computer Vision and Pattern Recognition*, 18-23 June 2018 2018, pp. 4510-4520, doi: 10.1109/CVPR.2018.00474.

[42] L. Nanni, S. Ghidoni, and S. Brahnam, "Deep features for training support vector machines," *Journal of Imaging,* vol. 7, no. 9, p. 177, 2021. [Online]. Available: https://www.mdpi.com/2313-433X/7/9/177.

[43] L. Nanni, E. D. Luca, and M. L. Facin, "Deep learning and hand-crafted features for virus image classification," *J. Imaging,* vol. 6, p. 143, 2020.

[44] A. R. Geus, A. R. Backes, and J. R. Souza, "Variability Evaluation of CNNs using Cross-validation on Viruses Images," in *VISIGRAPP*, University of Malta, Malta, 2020, pp. 626-632.

[45] Z.-j. Wen, Z. Liu, Y. Zong, and B. Li, "Latent Local Feature Extraction for Low-Resolution Virus Image Classification," *Journal of the Operations Research Society of China,* vol. 8, pp. 117-132, 2020.

[46] A. R. Backes and J. J. M. S. Junior, "Virus Classification by Using a Fusion of Texture Analysis Methods," *2020 International Conference on Systems, Signals and Image Processing (IWSSIP),* pp. 290-295, 2020.

[47] F. L. C. dos Santosa, M. Paci, L. Nanni, S. Brahnam, and J. Hyttinen, "Computer vision for virus image classification," *Biosystems Engineering,* vol. 138, no. October, pp. 11-22, 2015.

[48] L. Nanni, S. Ghidoni, and S. Brahnam, "Ensemble of Convolutional Neural Networks for Bioimage Classification," *Applied Computing and Informatics,* vol. 17, no. 1, pp. 19-35, 2021, doi: https://doi.org/10.1016/j.aci.2018.06.002.

[49] S. Boudra, I. Yahiaoui, and A. Behloul, "A set of statistical radial binary patterns for tree species identification based on bark images," *Multimedia Tools and Applications,* vol. 80, no. 15, pp. 22373-22404, 2021/06/01 2021, doi: 10.1007/s11042-020-08874-x.

[50] V. Remeš and M. Haindl, "Bark recognition using novel rotationally invariant multispectral textural features," *Pattern Recognit Lett,* vol. 125, pp. 612-617, 2019/07/01/ 2019, doi: https://doi.org/10.1016/j.patrec.2019.06.027.

[51] V. Remes and M. Haindl, "Rotationally Invariant Bark Recognition," in *Joint IAPR International Workshops on Statistical Techniques in Pattern Recognition (SPR) and Structural and Syntactic Pattern Recognition (SSPR S+SSPR)*, Beijing, China, 2018, pp. 22-31.

[52] F. Long, J.-J. Peng, W. Song, X. Xia, and J. Sang, "BloodCaps: A capsule network based model for the multiclassification of human peripheral blood cells," *Computer methods and programs in biomedicine,* vol. 202, no. Apr, p. 105972, 2021, doi: 10.1016/j.cmpb.2021.105972.

[53] F. Ucar, "Deep Learning Approach to Cell Classificatio in Human Peripheral Blood," in *2020 5th International Conference on Computer Science and Engineering (UBMK)*, 9-11 Sept. 2020, pp. 383-387, doi: 10.1109/UBMK50275.2020.9219480.

[54] Y. Song, W. Cai, H. Huang, D. Feng, Y. Wang, and M. Chen, "Bioimage classification with subcategory discriminant transform of high dimensional visual descriptors," *BMC Bioinformatics,* vol. 17, p. 465, 2016.

[55] L. P. Coelho *et al.*, "Determining the subcellular location of new proteins from microscope images using local features," *Bioinformatics,* vol. 29, no. 18, pp. 2343-2352, 2013.

[56] J. Zhou, S. Lamichhane, G. Sterne, B. Ye, and H. Peng, "BIOCAT: a pattern recognition platform for customizable biological image classification and annotation," *BMC Bioinformatics.,* vol. 14, p. 291, 2013.

[57] L. Shamir, N. Orlov, E. D. M., T. J. Macura, J. Johnston, and I. G. Goldberg, "Wndchrm - an open source utility for biological image analysis," *Source Code Biol Med,* vol. 3, no. 1, p. 13, 2008.

[58] S. Fort, H. Hu, and B. Lakshminarayanan, "Deep ensembles: A loss landscape perspective," *arXiv preprint arXiv:1912.02757,* 2019.